\pdfoutput=1
\documentclass[11pt]{article}

\usepackage[]{emnlp2021} % camara-ready

\usepackage{times}
\usepackage{latexsym}

\usepackage[T1]{fontenc}
\usepackage[utf8]{inputenc}

\usepackage{microtype}

\usepackage{mathtools}
\usepackage{amsmath}
\usepackage{multirow}
\usepackage{tabularx}
\usepackage{booktabs}
\usepackage{color}

\title{Exploring Decomposition for 
Table-based Fact Verification}

\author{
Xiaoyu Yang {\normalfont ~and~}
Xiaodan Zhu \\ 
  Ingenuity Labs Research Institute \& ECE, Queen's University, Canada \\ 
  \texttt{\{xiaoyu.yang, xiaodan.zhu\}@queensu.ca} \quad 
}

\begin{document}
\maketitle
\begin{abstract}
Fact verification based on structured data is challenging as it requires models to understand both natural language and symbolic operations performed over tables. Although pre-trained language models have demonstrated a strong capability in verifying simple statements, they struggle with complex statements that involve multiple operations.
In this paper, we improve fact verification by decomposing complex statements into simpler subproblems. 
Leveraging the programs synthesized by a weakly supervised semantic parser, we propose a program-guided approach to constructing a pseudo dataset for decomposition model training. 
The subproblems, together with their predicted answers, serve as the intermediate evidence to enhance our fact verification model.
Experiments show that our proposed approach achieves the new state-of-the-art performance, an 82.7\% accuracy, on the \textsc{TabFact} benchmark.
% Fact verification based on structured data is challenging as it requires models to understand both language understanding and symbolic operations.
% Although pre-trained transformers have demonstrated a strong capability to verify simple statements, while they struggle with handling complex statements that involve multiple operations. 
% In this paper, we propose to improve fact verification by decomposing complex statements into simpler subproblems.
% \xy{To avoid annotating gold decompositions,
% we propose a program-guided approach to create a pseudo dataset for decomposition model training. 
% We leverage the synthesized programs by parsing statements with a weakly supervised semantic parser.
% We solve the decomposed subproblems to obtain useful intermediate evidence for final verification predictions.}
% The experimental results show that our approach achieves new state-of-the-art results on the benchmark dataset \textsc{TABFACT}.

% After solving the decomposed subproblems to obtain useful intermediate evidence, which can be integrated into the verification decision-making process.
% To fully take advantage of simple samples, we also apply curriculum learning strategies to accelerate convergence and achieve more generalized results. 
% The results on \textsc{TabFact} show that this approach achieves state-of-the-art results.

\end{abstract}

\section{Introduction}
Fact verification aims to validate if a statement is entailed or refuted by given evidence. It has become crucial to many applications such as detecting fake news and rumor~\cite{rashkin2017truth, thorne2018fever, goodrich2019assessing, vaibhav2019sentence, kryscinski2019evaluating}.
While existing research mainly focuses on verification based on unstructured text~\cite{hanselowski2018ukp, yoneda2018ucl,liu2019finegrained, nie2019combining}, a recent trend is to explore structured data as evidence, which is ubiquitous in our daily life.

% figure-1
\begin{figure}[t]
  \centering
  \includegraphics
  [width=\columnwidth,trim={0cm 8.6cm 16.2cm 0cm},clip]
  {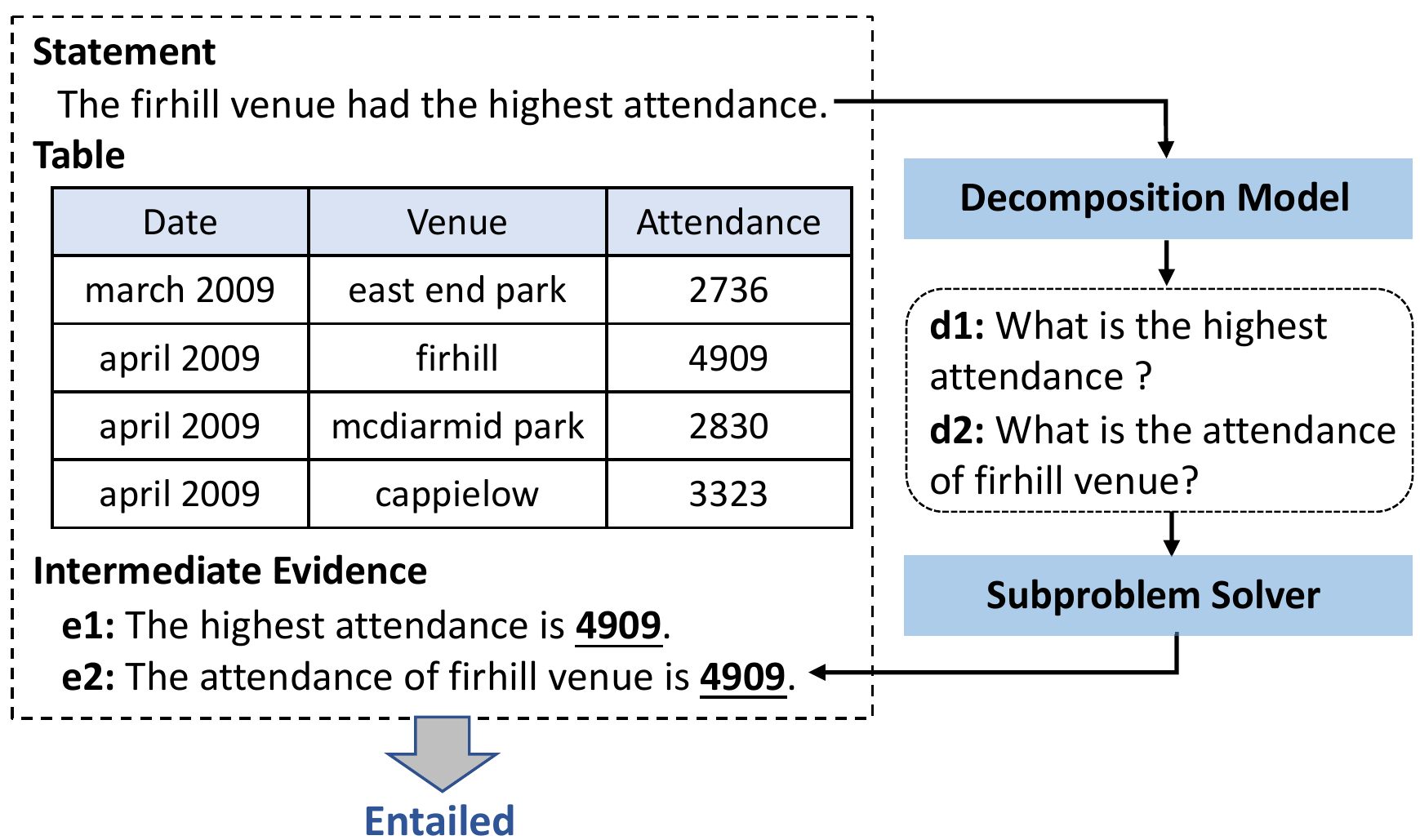}
  \caption{
  Overview of the proposed approach.
  An example of executable program parsed from the statement is: $eq\{max\{all\_rows;attendance\}; hop\{filter\_eq\\\{all\_rows;venue;firhill\};attendance\}\}$.}
\label{fig:intro_case}
\end{figure}

% % figure-1
% \begin{figure}[t]
%   \centering
%   \includegraphics[width=\columnwidth]{intro_fig_2.png}
% %   \caption{Overview: decompose the complex statement into simpler subproblems, solve the subproblems to obtain intermediate evidence sentences for fact verification.
% % An example of executable program parsed from the statement is: \textit{eq\{max\{all\_rows;attendance\}; hop\{filter\_eq\{all\_rows;venue;firhill\};attendance\}\}\}}.}
%   \caption{
%   Overview of the proposed approach.
%   An example of executable program parsed from the statement is: $eq\{max\{all\_rows;attendance\}; hop\{filter\_eq\\\{all\_rows;venue;firhill\};attendance\}\}\}$.}
% \label{fig:intro_case}
% \end{figure}

Verification performed with structured data
% , in addition to its real-life application values, 
presents research challenges of fundamental interests, as it involves both informal inference based on language understanding and symbolic operations such as mathematical operations (e.g., \texttt{count} and \texttt{max}). 
While all statements share the same set of operations, complex statements, which involve multiple operations, are more challenging than simple statements.
Pre-trained models such as BERT~\cite{devlin2018bert} have presented superior performances on verifying simple statements while still struggling with complex ones: a performance gap exists between the simple and complex tracks~\cite{chen2019tabfact}.
% The task inherently involves both simple verification cases that involve single operations each, or complex verification that consists of multiple operations used together to verify one statement. Given this difference, this paper investigates how to bring forward the performance of verification.

% We will first demonstrate that leveraging \textit{curriculum learning} to learn \textit{easy-to-difficult} benefits the task by \xy{improving the verification performance} but also making the training more stable and converge faster. 
% Our main focus in this paper, however, is on complex verification. We propose decomposition models to  convert complex statements into simpler subproblems and solve the decomposed problems to provide evidences for final verification decision, as shown in Figure\ref{fig:intro_case}. 

% In this paper, we propose decomposition models to convert complex statements into simpler subproblems and solve the decomposed problems to provide evidences for the final verification prediction as shown in Figure~\ref{fig:intro_case}.
In this paper, we propose to decompose complex statements into simpler subproblems to improve table-based fact verification, as shown in a simplified example in Figure~\ref{fig:intro_case}.
To avoid manually annotating gold decompositions, we design a program-guided pipeline to collect pseudo decompositions for training generation models by distinguishing four major decomposition types and designing templates accordingly.
% \xy{To avoid manually annotating gold decompositions,
% we design a program-guided pipeline to distinguish four major decomposition types and design templates accordingly to collect pseudo decompositions of statements for decomposition model training.}
% Since obtaining manual annotation for decomposed sentences is expensive, we design a program-guided pipeline to distinguish four major decomposition types and leverage templates to collect pseudo decomposition data and perform further data augmentation.
The programs we used are parsed from statements with a weakly supervised parser with the training signals from final verification labels. Figure~\ref{fig:intro_case} shows a statement-program example.
We adapt table-based natural language understanding systems to solve the decomposed subproblems.
After obtaining the answers to subproblems, we combine them in a pairwise manner as intermediate evidence to support the final prediction.

We perform experiments on the recently proposed benchmark~\textsc{TabFact}~\cite{chen2019tabfact} and achieve a new state-of-the-art performance, an 82.7\% accuracy. 
Further studies have been conducted to provide details on how the proposed models work.

\section{Method}
% We propose to verify complex compositional statements by decomposing them into simpler subproblems. Utilizing noisy programs derived from weakly-supervised semantic parsers~\cite{chen2019tabfact},  we create a pseudo-decomposition dataset, based on which we train a neural decomposition model (section~\ref{sec:decomposition}) that generates subproblems for each decomposable statement. After solving the subproblems (section~\ref{sec:solve}), we use the subproblems and its answers as evidence to enhance the final fact verification model (section~\ref{sec:combine evd}). 

\subsection{Task Formulation and Notations}
Given an evidence table $T$ and a statement $S$, we aim to predict  whether $T$ \textit{entails} or \textit{refutes} $S$, denoted by $y \in \{1, 0\}$.
% Given an evidence table $T$ and a statement $S$, we aim to predict  whether $T$ \textit{entails} or \textit{refutes} $S$, denoted as $y \in \{1, 0\}$.
% The evidence table $T = \{T_{i,j} |i \leq R , j \leq C \}$ has $R$ rows and $C$ columns, and $T_{i,j}$ refers to the cell value in $i$-th row, $j$-th column. 
For each statement $S$, the executable program derived from a semantic parser is denoted as $z$.
% For each statement $S$, the program $z$ derived from a semantic parser can be executed over table $T$. 
% The corresponding program $z$ of $S$ derived from semantic parser can be executed over the table $T$ for verification. 
An example of program is given in Figure~\ref{fig:intro_case}.
Each program $z=\{op_i\}_{i=1}^M$ consists of multiple symbolic operations $op_i$, and each operation contains an operator (e.g., \textit{max}) and arguments (e.g., \textit{all\_rows} and \textit{attendance}).
A complex statement $S$ can be decomposed into subproblems $D=\{d_i\}_{i=1}^{N}$, with the answers being $\{a_i\}_{i=1}^{N}$. 
Using combined problem-answer pairs as intermediate evidence $E=\{e_i\}_{i=1}^{N}$ where $e_i = (d_i, a_i)$, our model maximizes the objective ${\rm{log}}~p_{\theta}(y|T, S, E)$.

% To decompose the statement $S$, we utilize a collection of symbolic programs $z \in \mathcal{Z}$, which can be executed over the table $T$ to verify the statement. Each program consists of multiple executable symbolic operations (e.g., count, max). 

% We show an example statement-program pair in Figure~\ref{fig:example-1}.

% Given a statement $S$, a semantic consistent program $z$ can be executed over tables to yield a binary output for verification. 
% Each program $z=\{op_i\}_{i=1}^M$ consists of multiple executable symbolic operations.
% Figure~\ref{fig:example-1} gives an example of program.
% Our goal is to decompose a compositional statements $S$ into subproblems $\{s_i\}_{i=1}^{N}$, whose sub-answers are $\{a_i\}_{i=1}^{N}$. 
% Then we can get solved problem-answer pairs $\{e_i\}_{i=1}^{N}$ as intermediate evidence where $e_i = (s_i, a_i)$.
% When we integrate the intermediate evidence for final decision making process, we maximize the training objective as: ${\rm{log}} p(y|T, S, e_1, \dots, e_N)$.

\subsection{Statement Decomposition}
\label{sec:decomposition}

Constructing a high-quality dataset is key to the decomposition model training. Since semantic parsers can map statements into executable programs that not only capture the semantics but also reveal the compositional structures of the statements, we propose a program-guided pipeline to construct a pseudo decomposition dataset.

\subsubsection{Constructing Pseudo Decompositions}

\paragraph{Program Acquisition.}
Following~\citet{chen2019tabfact}, we use latent program algorithm (LPA) to parse each statement $S$ into a set of candidate programs $\mathcal{Z}=\{z_i\}_{i=1}^K$. 
To select the most semantically consistent program $z^*$ among all candidates and mitigate the impact of spurious programs, 
% i.e., incorrect programs execute to correct answers, 
we follow~\citet{yang2020program} to optimize the program selection model with a margin loss, which is detailed in Appendix~\ref{appendix:margin loss}.

By further removing programs that are label-inconsistent or cannot be split into two isolated sub-programs from the root operator, 
we obtain the remaining $(T,S,z)$ triples as the source of data construction\footnote{These triples do not involve any tables or statements in the dev/test set of the dataset used in this paper.}. 
% we obtain around 10k $(T,S,z)$ triples as the source of data construction\footnote{These triples do not involve any tables or statements in the dev/test set of the dataset used in this paper.}. 

\paragraph{Decomposition Templates.}
Programs are formal, unambiguous meaning representations for the corresponding statements. Designed to support automated inference, the program $z$ encodes the central feature of the statement $S$ and reveals its compositional structures.
Our statement decomposition is based on the structure of the program.
Specifically, we first extract program skeleton $z_s$ by omitting arguments in the selected program $z$, then we group the $(T,S,z)$ triples by $z_s$ to identify four major decomposition types: \textbf{conjunction\footnote{The conjunction type has overlap with the other three types in the cases that the sub-statements connected by conjunctions can be further decomposed.}}, \textbf{comparative}, \textbf{superlative}, and \textbf{uniqueness}.

\begin{figure}[t]
  \centering
  \includegraphics
  [width=\columnwidth,trim={0cm 0.5cm 15.9cm 0cm},clip]
  {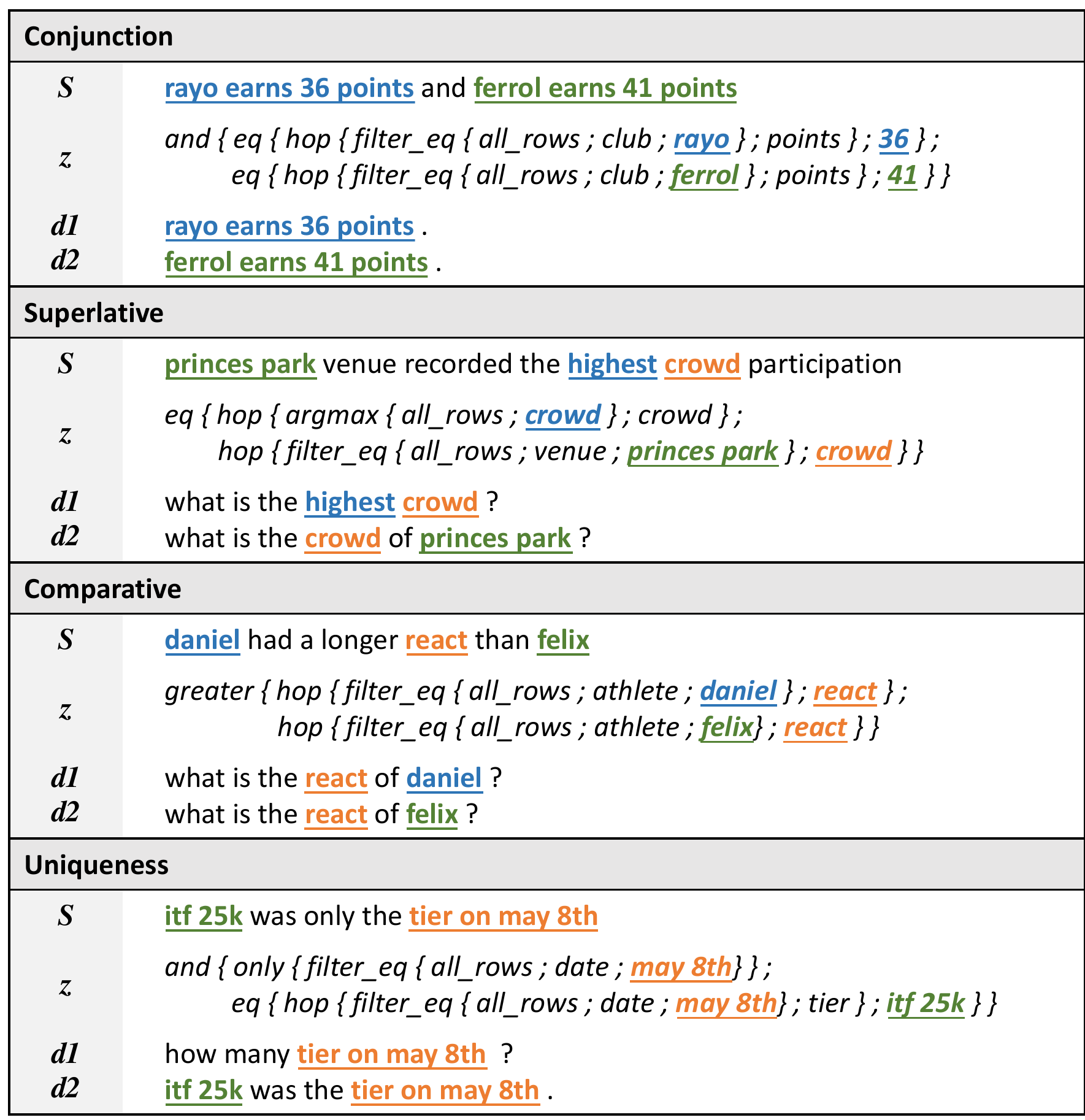}
  \caption{Decomposition templates.}
\label{fig:decomp_template}
\end{figure}

Some simple templates associated with each decomposition type are designed, which contain instructions on how to decompose the statement, and this manual process only takes a few hours.
In this way, we can construct pseudo decompositions, including sub-statements and sub-questions, by filling the slots in templates according to the original statements or program arguments.
Templates and decomposition examples can be found in Figure~\ref{fig:decomp_template}. 
% and we put more details in Appendix~\ref{}.
Each sample in our constructed pseudo dataset is denoted as a $(S,c,D')$ triple, where $c$ indicates one of the four types and $D'$ is a sequence of pseudo decompositions.
% We created a pseudo decomposition dataset in which each sample is $(S,c,D')$, where $c$ indicates one of the four types and $D'$ is a sequence of pseudo decompositions.

% \begin{figure}[t]
%   \centering
%   \includegraphics
% [width=\linewidth,trim={0cm 9cm 17.5cm 0cm},clip]
% {intro_case.pdf}
%   \caption{An example of fact verification over tables.}
% \label{fig:example-1}
% \end{figure}

% figure-2

% % figure-2
% \begin{figure}[t]
%   \centering
%   \includegraphics[width=\columnwidth]{Decomp_template.png}
%   \caption{Decomposition templates.}
% \label{fig:decomp_template}
% \end{figure}

\paragraph{Data Augmentation.}
With the $(T,S,z)$ triples, we perform data augmentation.
Since some entity mentions in $S$ and $z$ can be linked to cells in $T$, we can randomly replace the linked entities in $S$ and $z$ with different values in the same column of $T$. 
For example, in Figure~\ref{fig:intro_case}, we can replace the linked entity \textit{``firhill''} with another randomly selected entity \textit{``cappielow''}.
% in the same column.
Another augmentation strategy is inverting superlative and comparative. For the examples belong to \textbf{superlative} and \textbf{comparative}, we replace the original superlative or comparative in statements with its antonym, such as \textit{higher} $\rightarrow$ \textit{lower} and \textit{longest} $\rightarrow$ \textit{shortest}. 
% \xy{This will not effect our pseudo decomposition results.} 
% With data augmentation, we can not only enlarge the pseudo dataset, but also improve the quality of generation. 
% -- prev
% \xy{With data augmentation, we obtain 9,696 pseudo statement-decomposition pairs for generation model training.} 
In this way, we generate another 3k pseudo statement-decomposition pairs. In total, the final decomposition dataset used for generation model training includes 9,696 samples.
More statistics are available in Appendix~\ref{appendix:pseudo statistics}.

\subsubsection{Learning to Decompose}
% To make full use of decomposition type information, we design a two-step decomposition pipeline. 
% \xy{We first train a classifier to identify the decomposition type then involve the type information in training a decomposition model.}

\paragraph{Decomposition Type Detection.}
Given a statement $S$, we train a five-way classifier based on BERT to identify whether the statement is decomposable and if yes, which decomposition type it belongs to. 
In addition to the four types mentioned in the previous section, we add an \textit{atomic} category by involving additional non-decomposable samples. 
Only the statements not assigned with \textit{atomic} labels can be used for decomposition.
% Since the pseudo-decomposition dataset only involves four categories, we sample atomic samples from the simple track.
% The category detection can be seen as a pre-stage in statement decomposition, only the statements not assigned with ``atomic'' label are used for decomposition.

\paragraph{Decomposition Model.}
We finetune the GPT-2~\cite{radford2019language} on the pseudo dataset for decomposition generation.
Specifically, given the $(S, c, D')$ triple, we train the model by maximizing the likelihood $J={\rm log}~p_{\theta}(D'|S, c)$. 
We provide the model with gold decomposition type $c$ during training and the predicted type $\hat{c}$ during testing.
% We take greedy search as the decoding strategy.
Only informative and well-formed decompositions are involved in the subsequent process to enhance the downstream verification.
In case some sub-statements need further decomposition, it can be implemented by resending them to our pipeline\footnote{In most cases, there is no need to perform iterative decomposition, and we leave finer-grained decomposition for future research.}.

\subsection{Solving Subproblems}
\label{sec:solve}
We adapt~\textsc{TAPAS}~\cite{eisenschlos2020understanding}, a SOTA model on table-based fact verification and QA task,
to solve the decomposed subproblems.
Verifying sub-statements is formulated as a binary classification with the \textsc{TAPAS} model fine-tuned on the \textsc{TabFact}~\cite{chen2019tabfact} dataset.
To answer each sub-question, we use the \textsc{TAPAS} fine-tuned on WikiTableQuestions~\cite{pasupat2015compositional} dataset.
We combine the subproblems and their answers in a pairwise manner to obtain the intermediate evidence $E=\{e_i\}_{i=1}^N =\{(d_i, a_i)\}_{i=1}^N$, an example evidence is shown in Figure~\ref{fig:intro_case}. 

\subsection{Recombining Intermediate Evidence}
\label{sec:combine evd}
Downstream tasks can utilize the intermediate evidence in various ways. 
In this paper, we train a model to fuse the evidence $E$ together with the statement $S$ and table $T$ for table-based fact verification\footnote{For the non-decomposable statements, we put ``no evidence'' as the placeholder.}.
% There are various ways to use intermediate evidence, we train recomposition models to fuse the evidence together with original input (table and statement) to make final verification decisions.
Specifically, we jointly encode $S$ and $T$ with \textsc{TAPAS} to obtain the concentrated representation $\textbf{\textit{h}}_{ST}$.
% and take the embedding $\textbf{\textit{h}}_{ST}$ of the [CLS] token at the final layer as the concentrated representation.
We encode multiple evidence sentences with another \textsc{TAPAS} following the document-level encoder proposed in~\citet{liu2019text} by inserting [CLS] token at the beginning of every single sentence $e_i$ and taking the corresponding [CLS] embedding $\textbf{\textit{h}}_{e_i}$ in the final layer to represent $e_i$. 
% In addition, we encode multiple intermediate evidence sentences with another \textsc{TAPAS} following the document-level encoder proposed in~\cite{liu2019text}. 
% The representation of evidence $d_i$ is $\textbf{\textit{h}}_{d_i}$.

We employ a gated attention model to obtain aggregated evidence representation $\textbf{\textit{h}}_{evd}$ and predict the final label as follows:
\begin{equation}
\small
\centering
\begin{aligned}
    \textbf{\textit{h}}_{evd} = \sum_{i=0}^N{a_i \textbf{\textit{h}}_{e_i}}, \quad & a_i=\sigma(\textbf{\textit{h}}_{ST}^{T}\textbf{\textit{h}}_{e_i}) \\
    y = \sigma(W([\textbf{\textit{h}}_{evd}&\oplus \textbf{\textit{h}}_{ST}]))
    \nonumber
\end{aligned}
\end{equation}
% \begin{align}
%     \textbf{\textit{h}}_{evd} &= \sum_{i=0}^N{a_i \textbf{\textit{h}}_{e_i}}, a_i=\sigma(\textbf{\textit{h}}_{ST}^{\rm T}\textbf{\textit{h}}_{e_i}) 
%     \nonumber\\
%     y &= \sigma(W([\textbf{\textit{h}}_{evd}\oplus \textbf{\textit{h}}_{ST}]))
%     \nonumber
% \end{align}
where $W$ are trainable parameters, $\sigma$ is the sigmoid function, and $\oplus$ indicates concatenation.

\section{Experiments}
% \subsection{Setup}
\paragraph{Setup.}
% \paragraph{Data and Evaluation Metric.}
We conduct our experiments on a large-scale table-based fact verification benchmark \textsc{TabFact}~\cite{chen2019tabfact}.
The test set contains a \textit{simple} and \textit{complex} subset according to difficulty. 
A \textit{small} test set is further annotated with human performance. 
Following the previous work, we use \textit{accuracy} as the evaluation metric. 
Details of the data are listed in Appendix~\ref{appendix:data statistics}. 
% Implementation details of our models are in Appendix~\ref{appendix:implementation_details}. 
\paragraph{Implementation Details.}
% \label{appendix:implementation_details}
During fine-tuning the GPT-2 model to generate decomposition, we run the model with a batch size of 5 for 30 epochs using Adam optimizer~\cite{kingma2014adam} with a learning rate of 2e-6. 
We optimize the model for final verification prediction using Adam optimizer with a learning rate of 2e-5 and a batch size of 16. It usually takes 11 to 14 epochs to converge. Our code is available at~\url{https://github.com/arielsho/Decomposition-Table-Reasoning}.
% The models in our work are implemented with PyTorch~\cite{NEURIPS2019_9015}, and we will release the code for reproductivity.
% The models in our work are implemented with PyTorch~\cite{NEURIPS2019_9015}.

% Detailed data statistics are listed in Appendix~\ref{appendix:data statistics}.
% For model training, detailed hyperparameters are provided in Appendix~\ref{appendix:implementation_details} for reproducibility.

% \paragraph{Implementation Details.}
% During fine-tuning the GPT-2 model to generate decomposition, we run the model with a batch size of 5 for 30 epochs using Adam optimizer~\cite{kingma2014adam} with a learning rate of 2e-6. 
% We optimize the model for final verification prediction using Adam optimizer with a learning rate of 2e-5 and a batch size of 16. It usually takes 11 to 14 epochs to converge.

% ------------------------------------------------------------------
% table: Main table
\begin{table*}[ht]
\renewcommand{\arraystretch}{1.135}
\setlength{\tabcolsep}{5.5pt}
\small
\centering
\begin{minipage}{0.57\linewidth}
\centering
\small
\begin{tabular}{lccccc}
% \hline
\toprule[1pt]
Model  & Val  & Test & Simple & Complex & Small\\ \midrule
Human  & - & - & - & - & 92.1 \\ \midrule
LPA  & 57.7 & 58.2 & 68.5  & 53.2 & 61.5 \\
Table-BERT & 66.1 & 65.1 & 79.1  & 58.2 & 68.1 \\ 
LogicalFactChecker & 71.8 & 71.7 & 85.4 & 65.1 & 74.3 \\
HeterTFV & 72.5 & 72.3 & 85.9 & 65.7 & 74.2 \\
SAT & 73.3 & 73.2 & 85.5 & 67.2 & - \\
ProgVGAT & 74.9 & 74.4 & 88.3 & 67.6 & 76.2 \\
\textsc{TAPAS-Base} 
& 79.1 & 79.1 & 91.4 & 73.1 & 81.2 \\
\textsc{TAPAS-Large} 
& 81.5 & 81.2 & 93.0 & 75.5 & 84.1 \\
\midrule
\textbf{\textsc{Ours-Base}} 
& \textbf{80.8} & \textbf{80.7} & \textbf{91.9} & \textbf{75.1} & \textbf{82.5}
\\ 
\textbf{\textsc{Ours-Large}} 
& \textbf{82.7} & \textbf{82.7} & \textbf{93.6} & \textbf{77.4} & \textbf{84.7}
\\ 
% \hline
\bottomrule[1pt]
\end{tabular}
\centering
\caption{The accuracy (\%) of models on \textsc{TabFact}.}
\label{table:main results}
\end{minipage}\hfill
\renewcommand{\arraystretch}{1}
\begin{minipage}{0.42\linewidth}
\centering
\small
\begin{tabular}{lc|cc}
\toprule[1pt]
\multicolumn{2}{c|}{Type} 
& \textsc{Tapas-base} & \textsc{Ours-base}
\\ \midrule
Conj. &(15\%) & 79.9  & \textbf{82.6}     \\
Sup.  &(13\%) & 81.3  & \textbf{82.4}     \\
Comp. &(13\%) & 69.1  & \textbf{72.1}     \\
Uniq. &(\hphantom{.}6\hphantom{.}\%) & 70.4  & \textbf{74.4}     \\ 
Atomic &(53\%) & 81.7  & \textbf{82.5}     \\ 
\bottomrule[1pt]
\end{tabular}
\caption{Decompositions improve the performance on test set over 4 decomposition types.}
\label{table:detailed res}
\centering
\small
\begin{tabular}{l|cc}
\toprule[1pt]
            & BLEU-4 on Dev% & BLEU-1/2/3/4  
            & Human Val \\ \midrule
Our Decomp.     
% & \textbf{47.2/38.6/31.6/25.3}  
& \textbf{56.75}
% & \textbf{68}/100 
& \textbf{68\%}
\\
w/o data aug 
% & 46.2/37.7/30.8/24.7
& 48.42
% & 56/100  
& 56\%
\\ 
w/o type info
& 54.74
% & 63/100
& 63\%
\\
\bottomrule[1pt]
\end{tabular}
%\caption{BLEU score and human validation of decomposition results.}
\caption{Evaluation of decomposition quality.}
\label{ablation table: decomp-2}
\end{minipage}
\end{table*}
% \begin{minipage}{0.42\linewidth}
% \centering
% \small
% \begin{tabular}{lc|cc}
% \toprule[1pt]
% \multicolumn{2}{c|}{Type} & \begin{tabular}[c]{@{}c@{}}\textsc{TAPAS} \\ w/o decomp.\end{tabular} & \begin{tabular}[c]{@{}c@{}}Our Model\\ w/ decomp.\end{tabular} 
% \\ \midrule
% Conj &(31\%) & 79.9  & \textbf{82.6}     \\
% Sup  &(29\%) & 81.3  & \textbf{82.4}     \\
% Comp &(28\%) & 69.1  & \textbf{72.1}     \\
% Uniq &(12\%) & 70.4  & \textbf{74.4}     \\ 
% \bottomrule[1pt]
% \end{tabular}
% \caption{Decompositions improve the performance on the test set in \textsc{TabFact} on four categories (\textsc{Base} models).}
% \label{table:detailed res}
% \end{minipage}
% \end{table*}

% Decomposition - 2: eval decomposition generation 

% Decomposition - 1: decomposition rate in all data splits
\begin{table}[!ht]
\centering
\small
\begin{tabular}{l|ccccc}
\toprule[1pt]
            & train   & val  & test & simple  & complex  \\ \midrule
Our Decomp.      & \textbf{41.6} & \textbf{46.3} & \textbf{46.7} & \textbf{20.2} & \textbf{59.5}  \\
w/o data aug  & 35.2 & 39.1 & 39.4 & 16.3 & 50.7 \\ 
\bottomrule[1pt]
\end{tabular}
\caption{Percentage of valid decomposition on all splits in \textsc{TabFact}.}
%S-Test and C-Test refer to simple test and complex test, respectively.}
\label{ablation table: decomp-1}
\end{table}

\paragraph{Main Results.}
% Table~\ref{table:main results} presents the results of different models on \textsc{TabFact}.
% LPA is the baseline model proposed in~\citet{chen2019tabfact} which derives a program for each statement by ranking the synthesized program candidates and takes the program execution results as predictions.
% \textsc{TAPAS}~\cite{herzig2020tapas,eisenschlos2020understanding} is the previous SOTA model on \textsc{TabFact} which extends BERT's architecture to encode tables and is jointly pre-trained with text and tables.
% The model we built for the downstream task are based on \textsc{TAPAS}, and we present base and large versions correspond to the \textsc{TAPAS-Base} and \textsc{TAPAS-Large}, respectively. 
% Our base model achieves 80.7\% accuracy on the test set and our large model achieves 82.7\%. 
% The improvement on the complex test is 2.0\% and 1.9\%, respectively.
% \xy{Table~\ref{table:detailed res} shows detailed results of \textsc{TAPAS-BASE} and our base model across the four types of decomposition.}

% We compare our model with different baselines on \textsc{TabFact}:
% LPA is the baseline model proposed in~\citet{chen2019tabfact} which derives a program for each statement by ranking the synthesized program candidates and takes the program execution results as predictions;
% \textsc{TAPAS}~\cite{herzig2020tapas,eisenschlos2020understanding} is the previous SOTA model on \textsc{TabFact} which extends BERT's architecture to encode tables and is jointly pre-trained with text and tables.
We compare our model with different baselines on~\textsc{TabFact}, including LPA~\cite{chen2019tabfact}, Table-BERT~\cite{chen2019tabfact}, LogicalFactChecker~\cite{zhong2020logicalfactchecker}, HeterTFV~\cite{shi2020learn}, SAT~\cite{zhang2020table}, ProgVGAT~\cite{yang2020program}, and TAPAS~\cite{eisenschlos2020understanding}. Details of the compared systems can be found in Appendix~\ref{appandix:compared systems}.

Table~\ref{table:main results} presents the test accuracy of our \textsc{Base} model and \textsc{Large} model, which are built upon \textsc{TAPAS-Base} and \textsc{TAPAS-Large}, respectively. 
Results show that our model consistently outperforms the TAPAS baseline (80.7\% vs. 79.1\% for the base and 82.7\% vs. 81.2\% for the large model)\footnote{We also conduct significance tests over both the base and large models (the proposed model vs. TAPAS), with the one-tail t-test. For the base model, the p-value is 4.7e-6 and for the large model, 3.2e-7.}.  
We show in Table~\ref{table:detailed res} that our decomposition model decomposes roughly 47\% of the total \textsc{TabFact} test cases, and our model outperforms the TAPAS model over all types of decomposed statements.

\paragraph{Evaluation of Decompositions.}
We use both an automated metric and human validation to evaluate the decomposition quality. For the automated metric, we randomly sample 1,000 training cases from the pseudo decomposition dataset as the hold-out validation set, based on which we use BLEU-4~\cite{papineni2002bleu} to measure the generation quality. We also sample 100 decomposable cases from the \textsc{Tabfact} test set and ask three crowd workers to judge whether the model produces plausible decompositions.
The ablation results in Table~\ref{ablation table: decomp-2} indicate that data augmentation and the use of type information improve the decomposition quality, and the BLEU-4 score on the pseudo decomposition dataset well reflects the human judgements.

% The number of valid decompositions used by our final fact verification model varies according to the quality of decomposition, we remove defective decompositions to reduce noise in the downstream tasks. 
Since we remove the defective decompositions to reduce noise in the verification task, the number of decomposed cases involved by our final verification model varies according to the decomposition quality.
We provide the percentages of valid decompositions on all data splits of \textsc{TabFact} in Table~\ref{ablation table: decomp-1}.
The results show that our decompositions do not completely align with the simple/complex split provided in \textsc{TabFact}, and data augmentation can improve the number of valid decomposition by around 7\%. On the downstream verification task, a lower-quality decomposition ($39.4\%$) yields a $0.4\%$ performance drop compared to our proposed decomposition model ($46.7\%$).
\section{Related Work}
Existing work on fact verification is mainly based on evidences from unstructured text~\cite{thorne2018fever,hanselowski2018ukp, yoneda2018ucl,thorne2019fever2,nie2019combining,liu2019finegrained}.
Our work focuses on fact verification based on structured tables~\cite{chen2019tabfact}. Unlike the previous work~\cite{chen2019tabfact,zhong2020logicalfactchecker,shi2020learn,zhang2020table,yang2020program,eisenschlos2020understanding}, we propose a framework to verify statements via decomposition.

Sentence decomposition takes the form of Split-and-Rephrase proposed by~\citet{narayan2017split} to split a complex sentence into a sequence of shorter sentences while preserving original meanings~\cite{aharoni2018split,botha2018learning, guo2020fact}. 
In QA task, question decomposition has been applied to help answer multi-hop questions~\cite{iyyer2016answering,talmor2018web,min2019multi,wolfson2020break,perez2020unsupervised}.
Our work mainly focuses on decomposing statements for table-based fact verification with pseudo supervision from programs.

\section{Conclusion}
In this paper, we propose a framework to better verify the complex statements via decomposition. 
Without annotating gold decompositions, we propose a program-guided approach to creating pseudo decompositions on which we finetune the GPT-2 for decomposition generation.
By solving the decomposed subproblems,
% with table-based NLU systems, 
we can integrate useful intermediate evidence for final verification and improve the state-of-the-art performance to an 82.7\% accuracy on  \textsc{TabFact}.
% Future work will explore finer-grained decomposition of statements with more complex compositionality.

\section*{Acknowledgements}
We thank the anonymous reviewers for their insightful comments. We also thank Yufei Feng for his helpful comments and suggestions on the paper writing.
% need to mention Borealis AI
% \xy{This work was supported by Borealis AI through the Borealis AI Global Fellowship Award. (If you are acknowledging Borealis AI, please also email us to notify us.)}

\bibliography{anthology,custom}
\bibliographystyle{acl_natbib}

\clearpage
\appendix
\section{Appendix}
\label{sec:appendix}

\subsection{Program Selection}
\label{appendix:margin loss}
We fine-tune the BERT~\cite{devlin2018bert} to model $p_{\theta}(z|S)$, the probability of program $z$ being semantically consistent with $S$.
Since the gold programs are not available, 
we use the final verification labels as weak supervision. 
To mitigate the impact of spurious programs, i.e., programs execute to correct answers with incorrect operation combinations, 
we follow~\citet{yang2020program} to optimize the model with a margin loss: 
% $J=\max(p_{\theta}(z^{-}|S) -  p_{\theta}(z^{+}|S)+\gamma, 0)$, 
\begin{align}
    \small
    % J\!=\! \max\Big(p_{\theta}(z^{-}&|S) -  p_{\theta}(z^{+}|S)\!+\!\gamma, 0\Big)
    J\!=\! \max\Big(p(z^{-}&|S) -  p(z^{+}|S)\!+\!\gamma, 0\Big)
    \nonumber
    \label{eqn:margin_loss} 
\end{align}
where $z^-$ and $z^+$ denote the label-inconsistent and label-consistent programs with the highest probability, respectively. $\gamma$ is the parameter to control the margin. 
The margin loss can encourage selecting one program that is most semantically relevant to the statement while maintaining a margin between the positive (label-consistent) and the negative (label-inconsistent) programs.

\subsection{Statistics of Pesudo Dataset}
\label{appendix:pseudo statistics}
We have 9,696 pseudo statement-decomposition pairs in total, and the number of samples belong to four decomposition types is given in Table~\ref{table:pseudo_statistics}. 
To train the decomposition type detection model, we add an additional \textit{atomic} category with 1,739 statements.

\begin{table}[!htpb]
\centering
\begin{tabular}{c|cc}
\toprule[1pt]
\textbf{Decomp. Type} & \textbf{\# of samples}   \\ \midrule
Conjunctive  & 1,798            \\
Superlative  & 2,452            \\
Comparative  & 4,528            \\
Uniqueness   & 918            \\
\bottomrule[1pt]
\end{tabular}
\caption{Statistics of pseudo decomposition dataset.}
\label{table:pseudo_statistics}
\end{table}

\subsection{Statistics of \textsc{TabFact} Dataset}
\label{appendix:data statistics}
The statistics of \textsc{TabFact}~\cite{chen2019tabfact} can be found in Table~\ref{table:dataset_statistics}, a large-scale table-based fact verification benchmark dataset on which we evaluate our method.
The test set is further split into a \textit{simple} set and a \textit{complex} set, which include 4,171 and 8,608 sentences, respectively. A small test set with 1,998 samples are provided for human performance evaluation.

\begin{table}[!htpb]
\centering
\begin{tabular}{l|cccc}
\toprule[1pt]
\textbf{Split}         & \textbf{Sentence} & \textbf{Table} & \textbf{Row} & \textbf{Col} \\ \midrule
Train          & 92,283     & 13,182  & 14.1 & 5.5\\
Val            & 12,792     & \phantom{0}1,696   & 14.0 & 5.4\\
Test           & 12,779     & \phantom{0}1,695   & 14.2 & 5.4\\
\bottomrule[1pt]
\end{tabular}
\caption{Statistics of \textsc{\textsc{TabFact}}.}
\label{table:dataset_statistics}
\end{table}

% \subsection{Implementation Details}
% \label{appendix:implementation_details}
% During fine-tuning the GPT-2 model to generate decomposition, we run the model with a batch size of 5 for 30 epochs using Adam optimizer~\cite{kingma2014adam} with a learning rate of 2e-6. 
% We optimize the model for final verification prediction using Adam optimizer with a learning rate of 2e-5 and a batch size of 16. It usually takes 11 to 14 epochs to converge.
% % The models in our work are implemented with PyTorch~\cite{NEURIPS2019_9015}, and we will release the code for reproductivity.
% The models in our work are implemented with PyTorch~\cite{NEURIPS2019_9015}.
% To accelerate model convergence and achieve more stable performance, we train the final model with a curriculum learning strategy~\cite{bengio2009curriculum}. 
% Specifically, the model starts by training to verify simple statements for the first 10\% to 15\% of the total training steps, then we add complex track samples to the training set for the remaining steps.

\subsection{Compared Systems}
\label{appandix:compared systems}
\begin{itemize}
  \item \textbf{LPA}~\cite{chen2019tabfact} derives a program for each statement by ranking the synthesized program candidates and takes the program execution results as predictions.
  \item \textbf{Table-BERT}~\cite{chen2019tabfact} takes a linearized table and a statement as the input of BERT for fact verification.
  \item \textbf{LogicalFactChecker}~\cite{zhong2020logicalfactchecker} utilizes the structures of programs to prune irrelevant information in tables and modularize symbolic operations with module networks. 
  \item \textbf{HeterTFV}~\cite{shi2020learn} is a graph-based reasoning approach to combining linguistic information and symbolic information.
  \item \textbf{SAT}~\cite{zhang2020table} is a structure-aware Transformer that encodes structured tables by injecting the structural information into the mask of the self-attention layer.
  \item \textbf{ProgVGAT}~\cite{yang2020program} leverages the symbolic operation information to enhance verification with a verbalization technique and a graph-based network.
  \item \textbf{TAPAS}~\cite{herzig2020tapas,eisenschlos2020understanding} is the previous SOTA model on \textsc{TabFact} which extends BERT's architecture to encode tables and is jointly pre-trained with text and tables.
  
\end{itemize}

% \paragraph{Evaluation of Subproblem Solving.}
% We utilize \textsc{TAPAS} to solve the decomposed subproblems. 
% Although there is a performance gap between \textsc{TAPAS-Base} and \textsc{TAPAS-large} on \textsc{TabFact} (79.1\% vs. 81.2\% acc) and WTQ (46.4\% vs. 51.0\% acc), the two models have similar performance in solving the decomposed subproblems, thus result similar accuracy in the downstream task.

\end{document}